\documentclass[letterpaper, 10 pt, conference]{ieeeconf}  

\IEEEoverridecommandlockouts                              

\usepackage{graphics} 
\usepackage{epsfig} 
\usepackage{times} 
\usepackage{amsmath} 
\usepackage{amssymb}  
\usepackage{multirow}
\usepackage{comment}
\usepackage{caption}
\usepackage{subcaption}
\usepackage{booktabs}       
\usepackage{subcaption}
\usepackage{float}
\usepackage{xcolor}
\usepackage{color,soul}
\usepackage[pagebackref=true,breaklinks=true,colorlinks,bookmarks=false]{hyperref}
\DeclareCaptionLabelFormat{andtable}{#1~#2  \&  \tablename~\thetable}

\title{\LARGE \bf
Evaluating Computer Vision Techniques for Urban Mobility on Large-Scale, Unconstrained Roads
}

\author{
$^{*}$Harish Rithish$^{1}$, $^{*}$Raghava Modhugu$^{1}$, $^{*}$Ranjith Reddy$^{1}$, $^{*}$Rohit Saluja$^{1}$, C.V. Jawahar$^{1}$ 
\thanks{*Authors contributed equally to this research.}
\thanks{$^{1}$The authors are with Center for Visual Information Technology (CVIT), IIIT Hyderabad, India.
        {\tt\small \{harishrithish7,ranjithreddy1061995\}@gmail.com},
        {\tt\small \{durga.nagendra, rohit.saluja\}@reasearch.iiit.}
        {\tt\small ac.in, jawahar@iiit.ac.in}}%
}

\begin{document}

\maketitle
\thispagestyle{empty}
\pagestyle{empty}

\begin{abstract}
Conventional approaches for addressing road safety rely on manual interventions or immobile CCTV infrastructure.  
Such methods are expensive in enforcing compliance to traffic rules and do not scale to large road networks. 
This paper proposes a simple mobile imaging setup to address several common problems in road safety at scale.
We use recent computer vision techniques to identify possible irregularities on roads, the absence of street lights, and defective traffic signs using videos from a moving camera-mounted vehicle.
Beyond the inspection of static road infrastructure, we also demonstrate the mobile imaging solution's applicability to spot traffic violations.
Before deploying our system in the real-world, we investigate the strengths and shortcomings of computer vision techniques on thirteen condition-based hierarchical labels.
These conditions include different timings, road type, traffic density, and state of road damage.
Our demonstrations are then carried out on 2000 km of unconstrained road scenes, captured across an entire city.
Through this, we quantitatively measure the overall safety of roads in the city through carefully constructed metrics.
We also show an interactive dashboard for visually inspecting and initiating action in a time, labor and cost-efficient manner.
Code, models, and datasets used in this work will be publicly released.

\end{abstract}

\section{Introduction}~\label{Intro}
The ever-increasing urban population on roads has
increased the number of accidents and deaths~\cite{tran2017collaborative}, many of which could have been avoided with the effective use of technology. 
The escalating traffic also poses technical challenges 
in the urban mobility-related tasks to (i) driver assistance, (ii) reliable audit and maintenance of the infrastructure,  and (iii) detection and prevention of traffic violations. This is especially true for unstructured driving situations and unconstrained roads, which are common in developing countries.


\begin{figure}[t]
 \begin{center}
 \frame{\includegraphics[width=\linewidth]{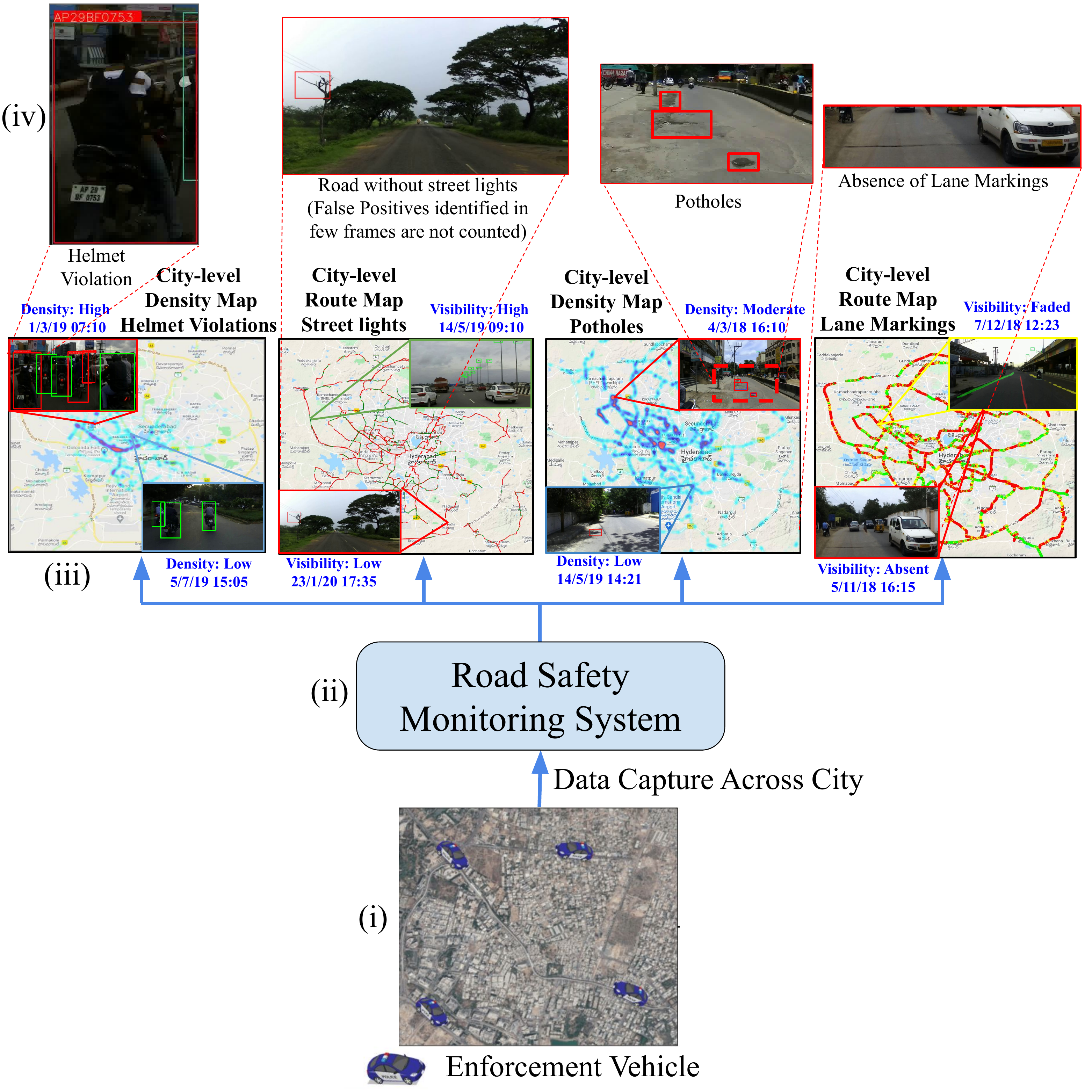}}
 \end{center}
   \caption{Our pipeline to identify irregularities on unconstrained roads. Bottom-to-top: i) Data capture via mobile setup, ii) Monitoring system to identify irregularities in data, iii) City-level maps linked to evidence of irregularities, and iv) left to right: sample helmet violation, road without street light, road with potholes, and absence of lane markings.}
 \label{fig:traffic_irregularities}
\end{figure}

Unconstrained roads suffer from uneven quality of infrastructures, such as low lighting, inadequate signage, and damaged roads, making road safety a challenging problem~\cite{acm_comm}. 
Fig.~\ref{fig:traffic_irregularities} (top) gives a glimpse of the potential problems seen in an unstructured situation. These include (i) traffic violations, (ii) missing, rusted or non-functional-signage, street lights, and traffic lights, (iii) uneven/partly damaged roads, (iv) illegal movement patterns. 
Therefore, tasks such as regular audit and maintenance of road infrastructure and traffic violation monitoring have become critical for safety.

Current approaches for addressing road safety, however, rely on manual interventions or large camera networks. 
This is not scalable or economical for massive road lengths in ever-expanding urban settlements~\cite{ma2017large, Qian_2020_CVPR_Workshops}.
Thus, we propose to use a simple mobile imaging system to address several common problems in road safety at scale using recent computer vision algorithms. As shown in Fig.~\ref{fig:traffic_irregularities}, the mobile imaging system captures the data across the city, and the proposed road safety system is used to generate the city-level maps for different irregularities.
The proposed system is highly automated, affordable, and scalable.
We demonstrate our system on 2000 km of unconstrained road scenes, captured across an entire city.
Through this, we quantitatively assess the overall safety of roads in the city through carefully constructed metrics.
We also develop an interactive dashboard for visually inspecting and initiating action in a time, labor, and cost-efficient manner.

\begin{table*}[!ht]
\caption{$2000$ km of video data is classified into following categories in different conditions at a resolution of 1 second. Criterion to classify in each condition is provided in the second row. The condition to determine a category is provided within the brackets. \textit{v} denotes visual discretion.}
\label{tab:data-categories}
\centering
\resizebox{0.7\textwidth}{!}{%
\def\arraystretch{1.2}%
\begin{tabular}{|l|l|l|l|l|}
\hline
Condition & Road Type & Traffic Density & Road Damage & Time \\ 
\hline
Criterion & \textit{\#Lanes} & \textit{\#Vehicles} & \textit{\#Potholes}  & \textit{Capture time (24 hr)} \\ 
\hline
Categories & Narrow (1) & Sparse (\textless{}=4) & Low (\textless{}=2) & Morning (7-12) \\
 & Standard (2-3) & Moderate (5-8) & Moderate (3-4) & Noon (12-16) \\
 & Highway (\textgreater{}3) & Dense (\textgreater{}8) & High (\textgreater{}4) & Evening (16-19) \\
 & Bridge (\textit{v}) & & & \\
\hline
\end{tabular}%
}
\end{table*}

However, vision-based detection algorithms in unstructured settings can be affected by continuously changing backgrounds, different road types (shape, color), variable lighting conditions, and weather.
Therefore, we investigate the behavior of computer vision algorithms in diverse settings before deploying them in the real-world.
For this purpose, we collect data in unconstrained road scenes and annotate them with thirteen condition-based hierarchical labels.
These conditions include different timings, road type, traffic density, and state of road damage.
We study the strengths and shortcomings of recent computer vision techniques in identifying irregularities in road infrastructure and spotting traffic violations on diverse and unconstrained roads through this evaluation.


We select tasks that are critical for road safety and reflect the overall compliance to traffic rules in a city.
As part of road infrastructure inspection, we identify possible irregularities in streets (including missing lane markings and potholes), absence of street lights, and defective traffic signs.
Such infrastructure provides essential visual cues for navigation and ensures safe mobility.
In the case of traffic violations, we focus on two-wheeled motor vehicle riders, as they account for nearly $30\%$ of road accident deaths worldwide~\cite{WHOReport}. 
Among them, around $74\%$ of riders were found to be not wearing protective helmets~\cite{AccidentsReport}. 
Therefore, we focus on spotting helmet violations.


To summarize, the main contributions of this paper are:
\begin{itemize}
    \item We investigate the behavior of recent computer vision techniques in inspecting road infrastructure and traffic violations under diverse and unconstrained conditions. 
    \item We demonstrate our system at a city-scale and assess the safety of roads.
    \item We release a 75-hour dataset of long video sequences captured in unstructured settings. These sequences are classified into 13 different conditions of time, road type, state of road damage, and traffic.
\end{itemize}







\section{Related Work}~\label{sec_related_work}
Many previous works have demonstrated the applications of computer vision models for object detection, semantic segmentation, and vehicle tracking in urban mobility~\cite{ovsep2018track,sudhir2018cityscape, bochkovskiy2020yolov4, Qian_2020_CVPR_Workshops}. 
The challenges like Nvidia AI city\cite{Naphade_2018_CVPR_Workshops}, CityFlow\cite{cityflow2018}, and AutoNUE\cite{Autonue2018} have popularized city-scale mobility-related tasks.

\subsection{Road Safety Systems}
Early work on road safety \cite{cindy2001object, petersson2003driver} demonstrated the effective collaboration between perception and control for
increasing safety. Pedestrian detection~\cite{geronimo2009survey}, driver state and behavior characterization, etc., became an integral part of ADAS system over the years. Zhang et al.~\cite{zhang2018dijkstra} demonstrated a stereo vision-based training-free approach to detect road regions accurately and robustly. Comprehensive surveys on vision-based traffic monitoring and vision applications in urban analytics are conducted regularly ~\cite{datondji2016survey, jain2019review, ibrahim2020understanding}.

Dynamic objects around a vehicle also impact the safety and performance of the vehicles.
Billones et al. ~\cite{billones2016intelligent} measure a vehicle's speed by counting the frames it takes to pass a region of interest or reference lines.
Singh et al.~\cite{Singh2016Traffic} present a framework for visual big data analytics that automatically detects bike-riders without helmets. They further discuss the challenges associated with city-scale surveillance for traffic control. Recent works have shown CNN to produce better results for helmet detection and classification~\cite{vishnu2017detection}. Frossard et al.~\cite{frossard2019deepsignals} detect driver's intent by detecting turn signals and emergency flashers in video sequences. Datasets and robust models on the pedestrian intent prediction are also present in the literature~\cite{kitani2012activity, keller2013will, kooij2014context, rasouli2017they, liu2020spatiotemporal}. 
In comparison to these works, we use recent vision techniques (refer Section~\ref{sec:system}) to inspect helmet violations at scale on the unconstrained road scenes under different conditions.

\begin{figure*}[!ht]
    \centering
    \includegraphics[width =0.8\linewidth,  ]{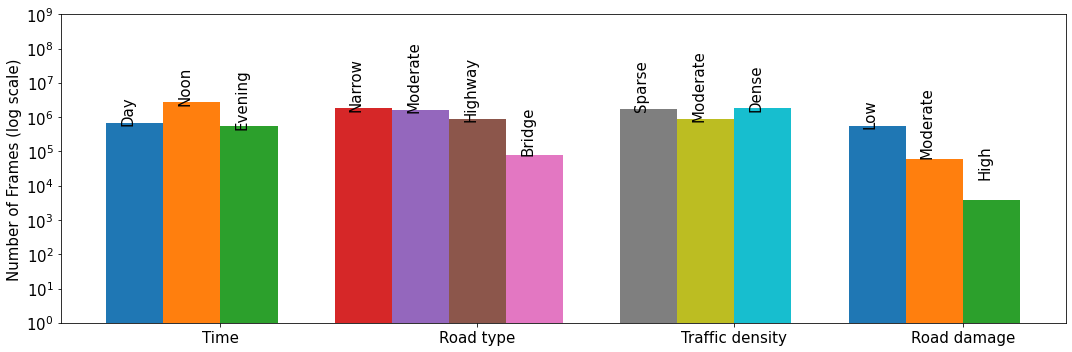}
    \caption{
     Distribution of the 13 condition-based  hierarchical frame-level labels. These conditions represent diversity in road type, traffic density, time of capture, and state of road damage. The annotations are carried out 2000 km of road scenes captured in unstructured settings.
    }
    \label{fig:dataset_summary}
\end{figure*}

\subsection{Road Inspection Systems} 
The soundness and suitable perceptibility of road infrastructure form the essential elements of driver safety. 
Burschka et al.~\cite{burschka2005vision} introduce a vision-based system for traffic sign detection and ego-motion estimation. Another early work involves detecting cracked regions using a weakly supervised superpixel classifier trained on $220$ images~\cite{varadharajan2014vision}.

To reduce asphalt pavement distresses, Kanza et al.~\cite{azhar2016computer} detect and localize pothole in $120$ cropped road images using histogram of oriented gradients (HOG) features and Naïve Bayes classifier. Zhang et al.~\cite{zhang2016road} employ Convolutional Neural Network (CNN) to classify image patches of the road as defective or non-defective and evaluate it on $500$ images. Yerram et al.~\cite{sudhir2018cityscape} train a multi-step ERFNet based model on around $1000$ unconstrained road scenes to segment road pixels into $9$ different kinds of defects. However, all these road inspection methods are applied at a small scale compared to the approach in this paper. 

The work closest to ours in this direction are 
i) Ma et al.~\cite{ma2017large} who utilize Fisher vectors with Convolutional Neural Networks (CNN) to classify $700$K images from $70$K street segments into poor, fair, and good, and ii) Modhugu et al.~\cite{RoadInfra_NCVPRIPG} who demonstrates an automatic model for road infrastructure audit concerning traffic signs, street lights, and lane markings. In contrast to these works, we extend the range of functionalities, volume of training data and test the system's scalability extensively on a larger dataset spanning $2000$~km of unconstrained roads. %

\section{Dataset}


\begin{figure}[t]
    \begin{subfigure}{.2367\textwidth}
      \centering
      \includegraphics[width=\linewidth]{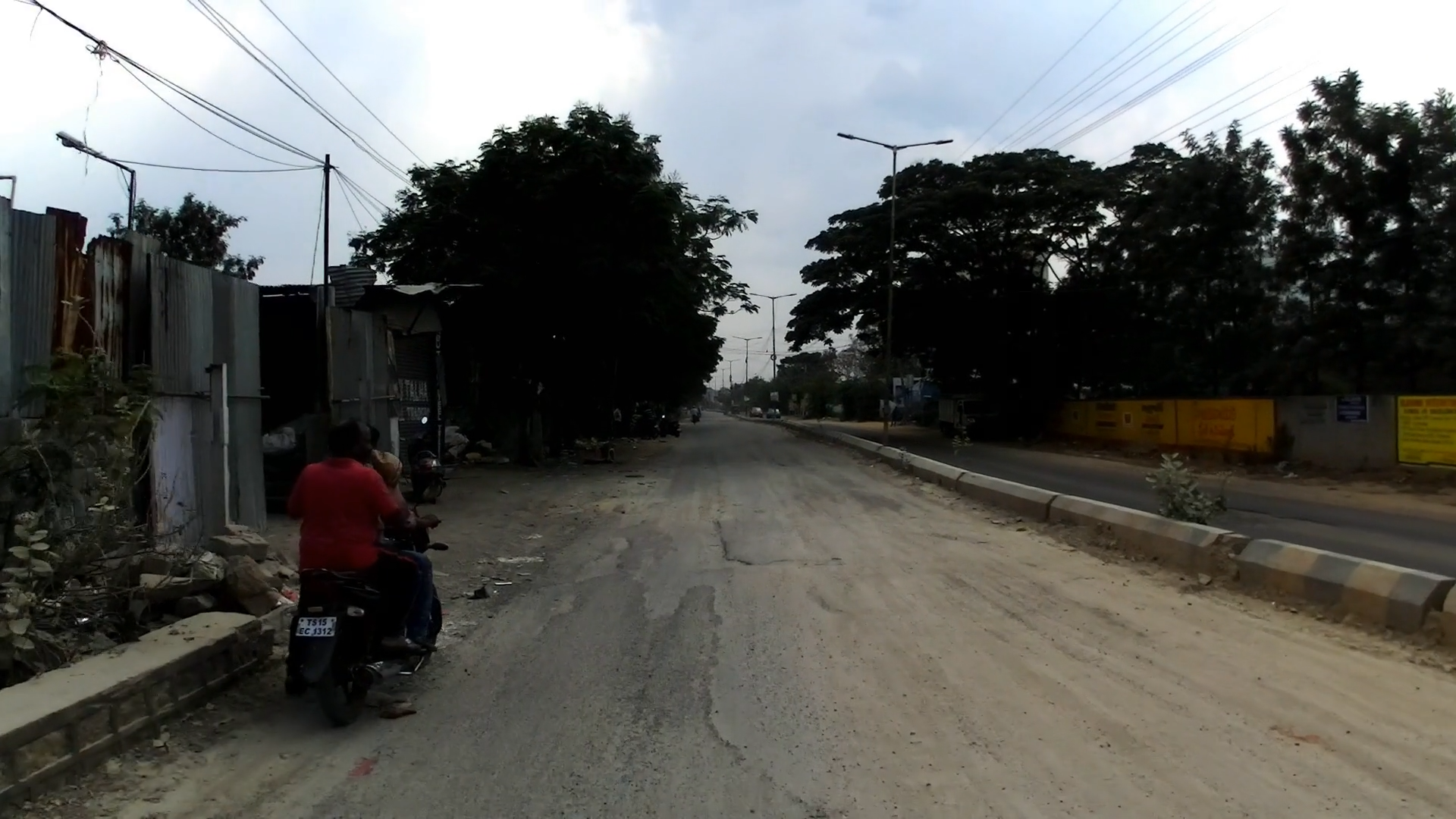}  
    \end{subfigure}
    \begin{subfigure}{.2367\textwidth}
      \centering
      \includegraphics[width=\linewidth]{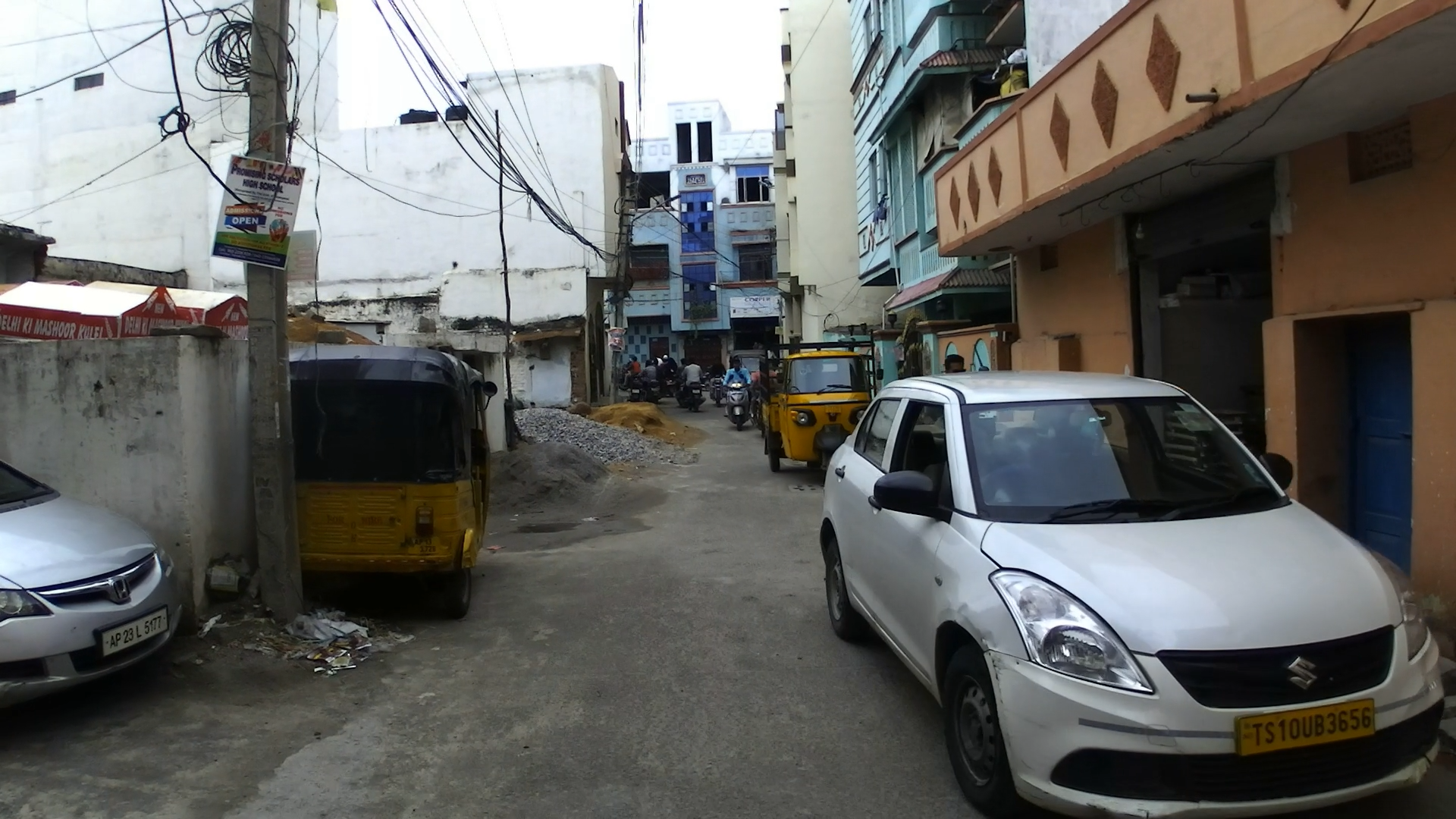}  
    \end{subfigure}\\\vspace{1.2mm}\\
    \begin{subfigure}{.2367\textwidth}
      \centering
      \includegraphics[trim={25cm 11.26cm 15cm 11.26cm},clip,width=\linewidth]{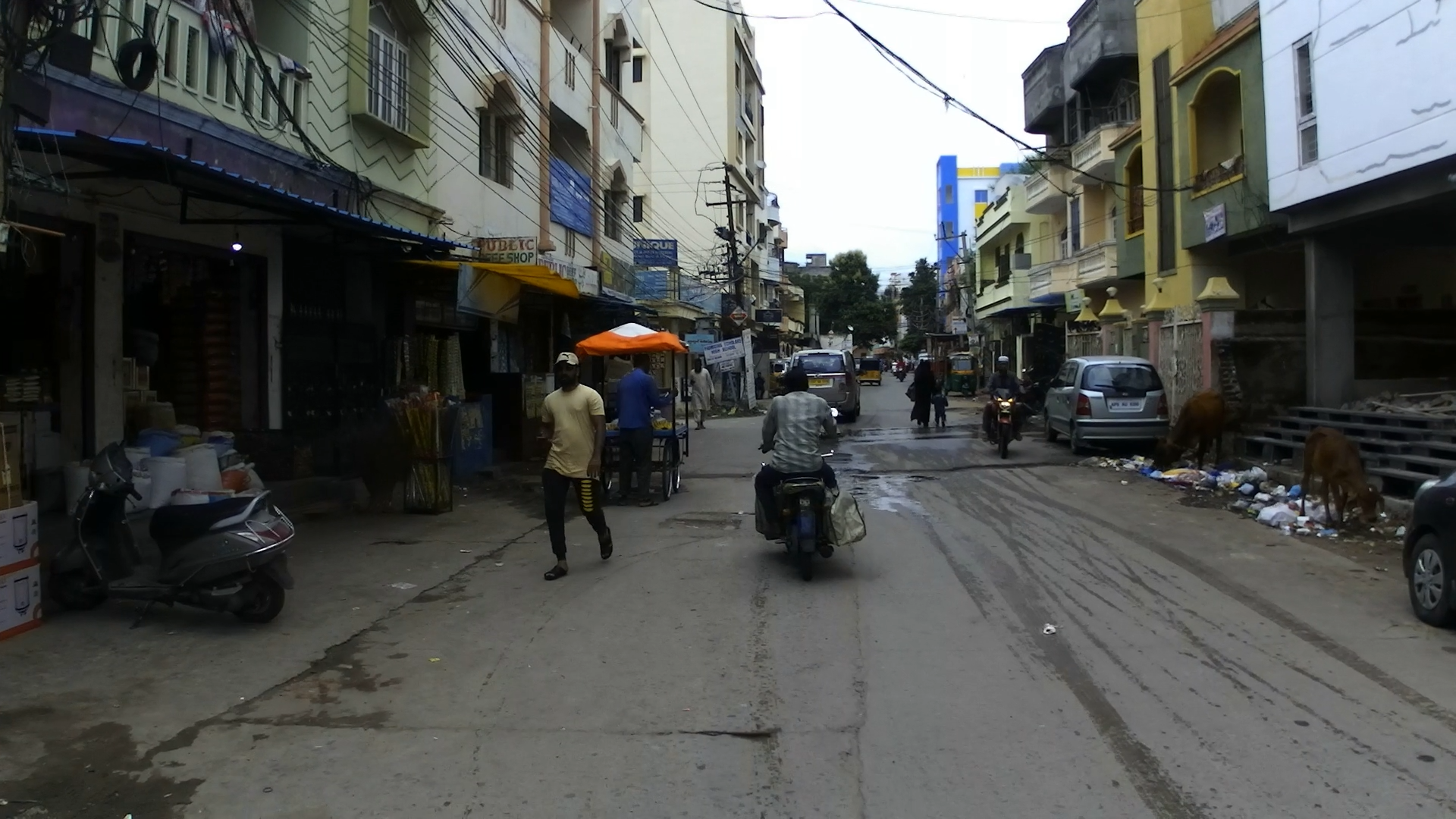}
    \end{subfigure}
    \begin{subfigure}{.2367\textwidth}
      \centering
      \includegraphics[width=\linewidth]{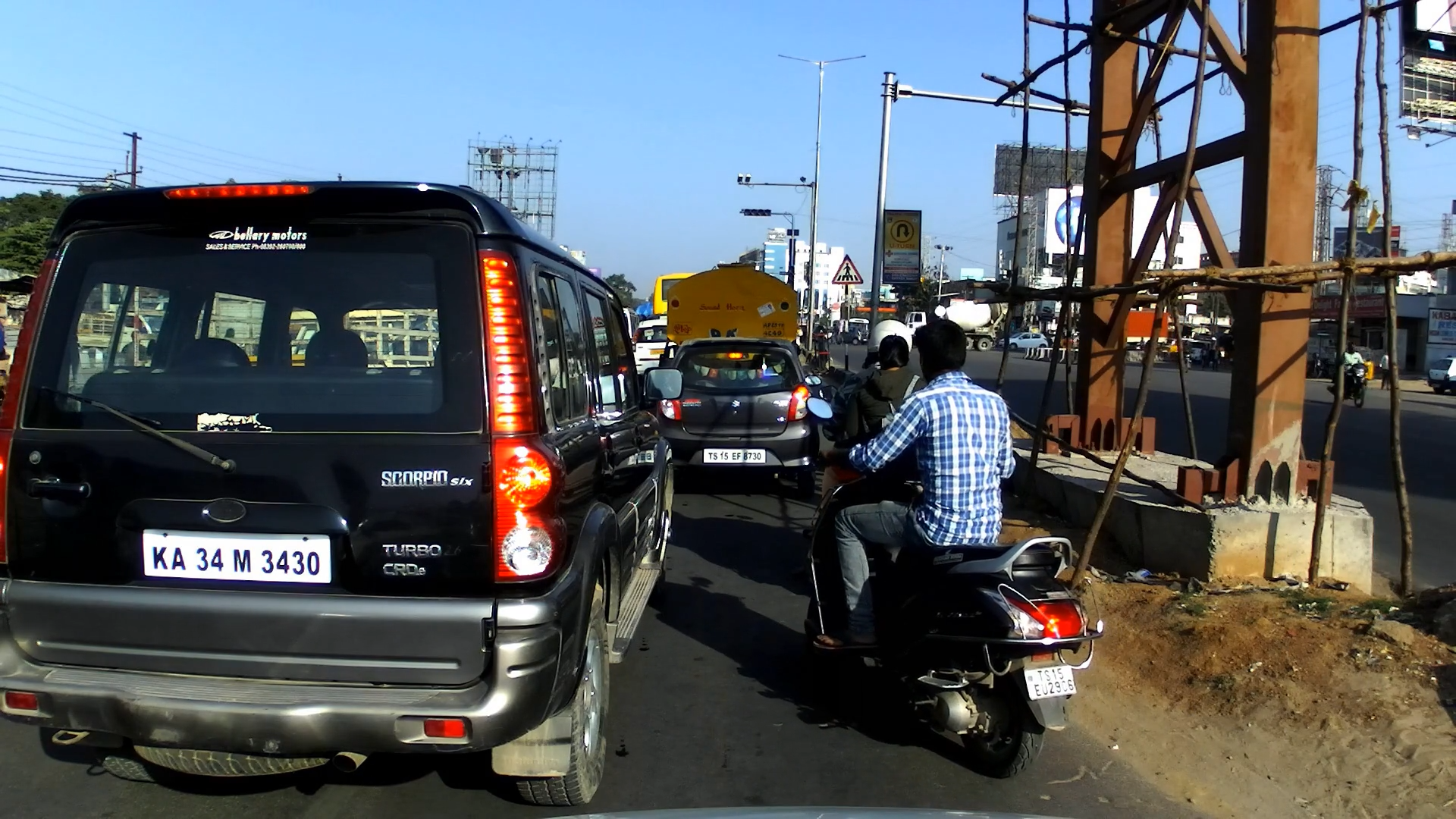}
    \end{subfigure}
    \caption{Dangerous road conditions. Clockwise from top-left: bumpy and muddy road, low-hanging wires hazardous for heavy vehicles,unguarded under construction structure and narrow street with two-way traffic.}
    \label{fig:capture_road}
\end{figure}

\begin{figure}[t]
    \begin{subfigure}{.2367\textwidth}
      \centering
      \includegraphics[width=\linewidth,]{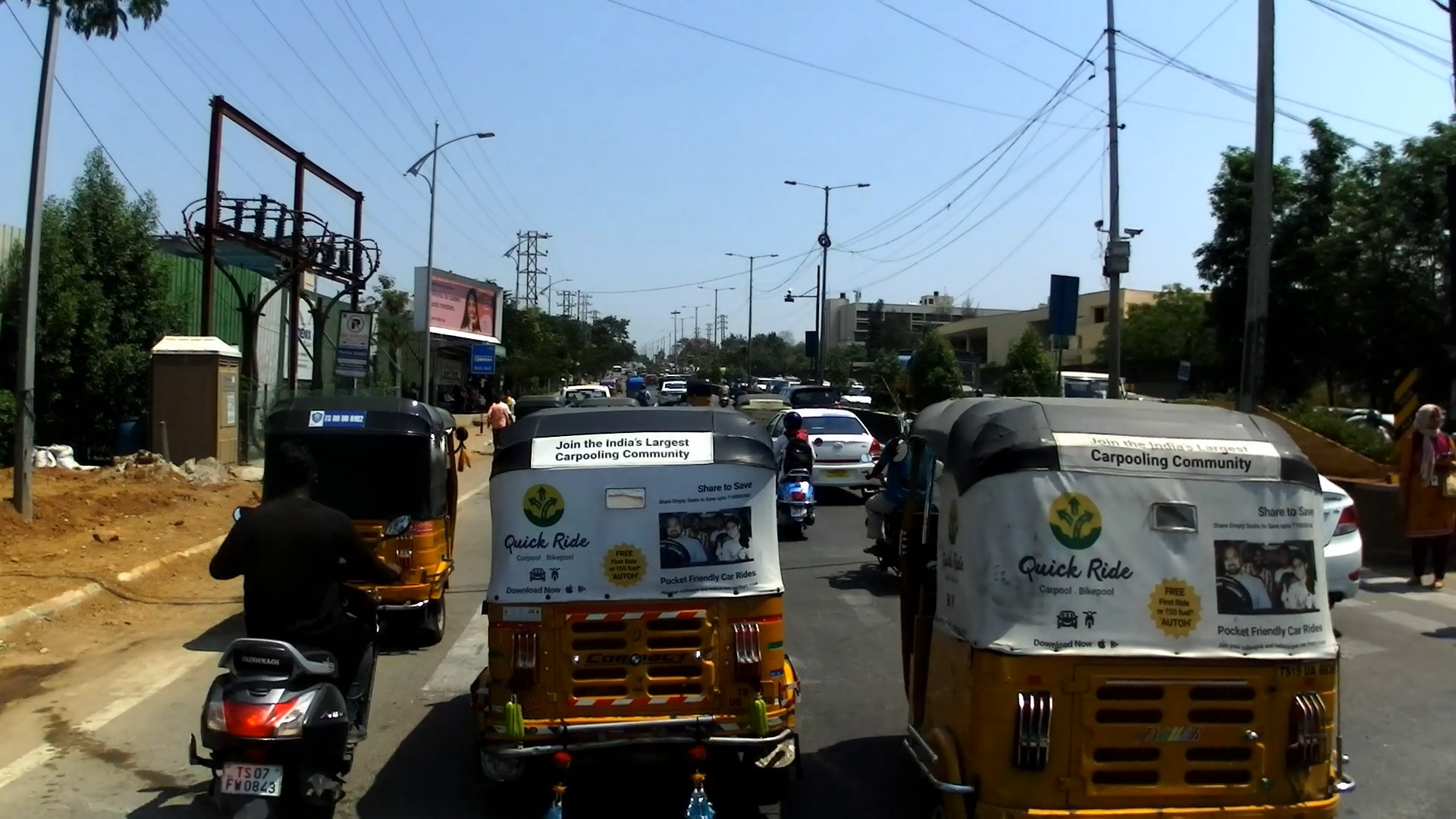}  
    \end{subfigure}
    \begin{subfigure}{.2367\textwidth}
      \centering
      \includegraphics[width=\linewidth,]{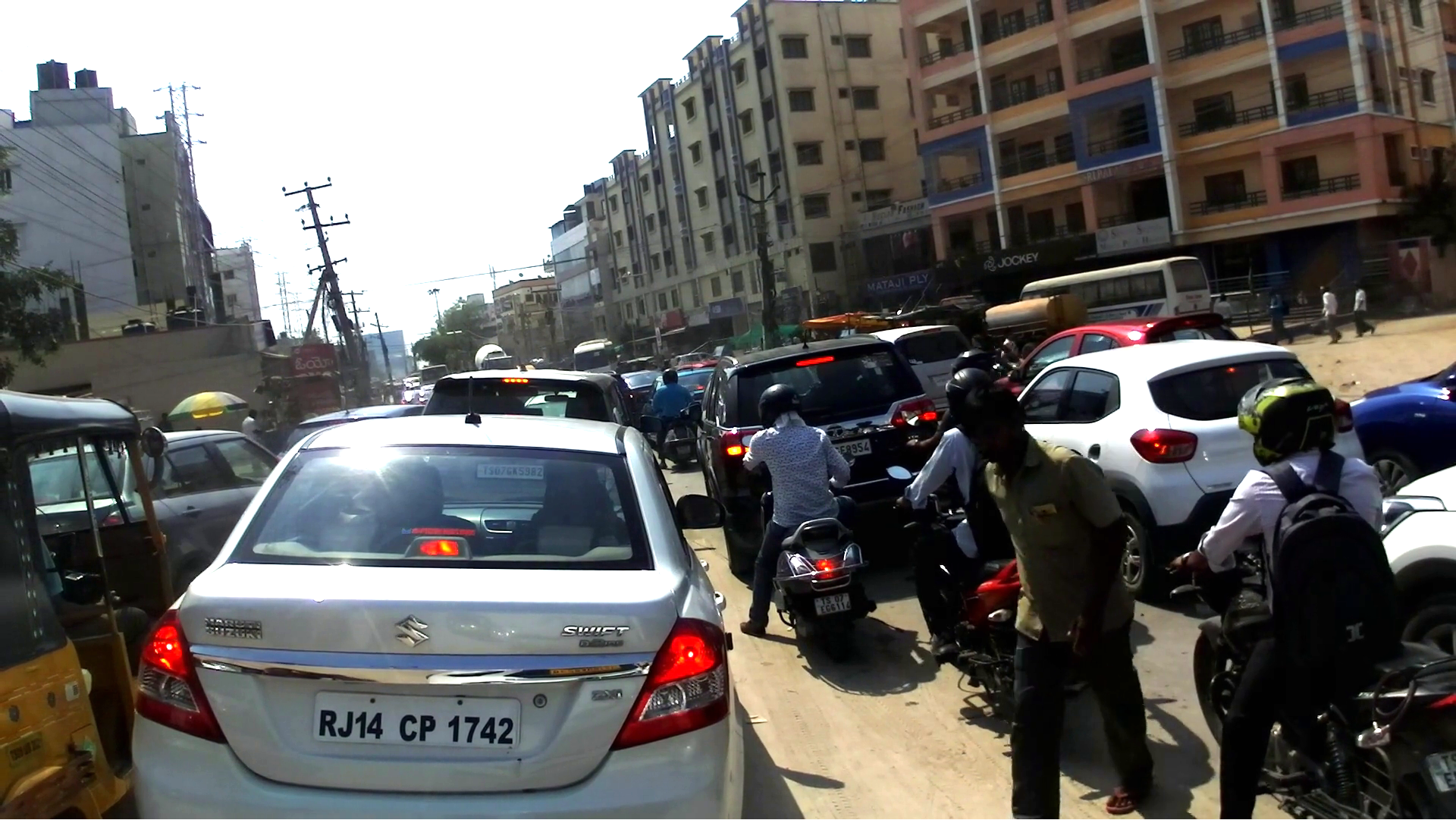} 
    \end{subfigure}\\\vspace{1.2mm}\\
    \begin{subfigure}{.2367\textwidth}
      \centering
      \includegraphics[width=\linewidth, ]{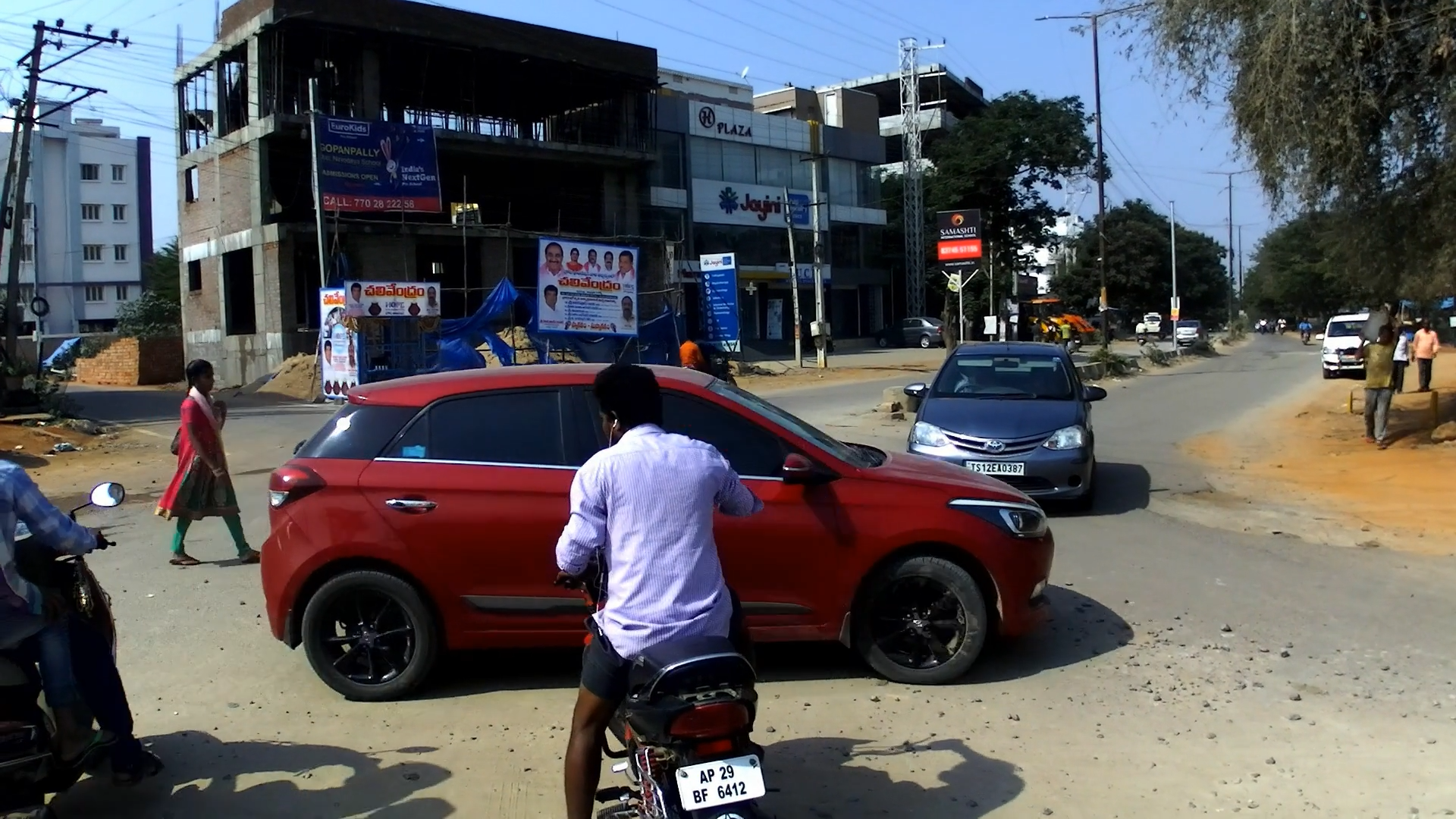}
    \end{subfigure}
    \begin{subfigure}{.2367\textwidth}
      \centering                   \includegraphics[width=\linewidth,]{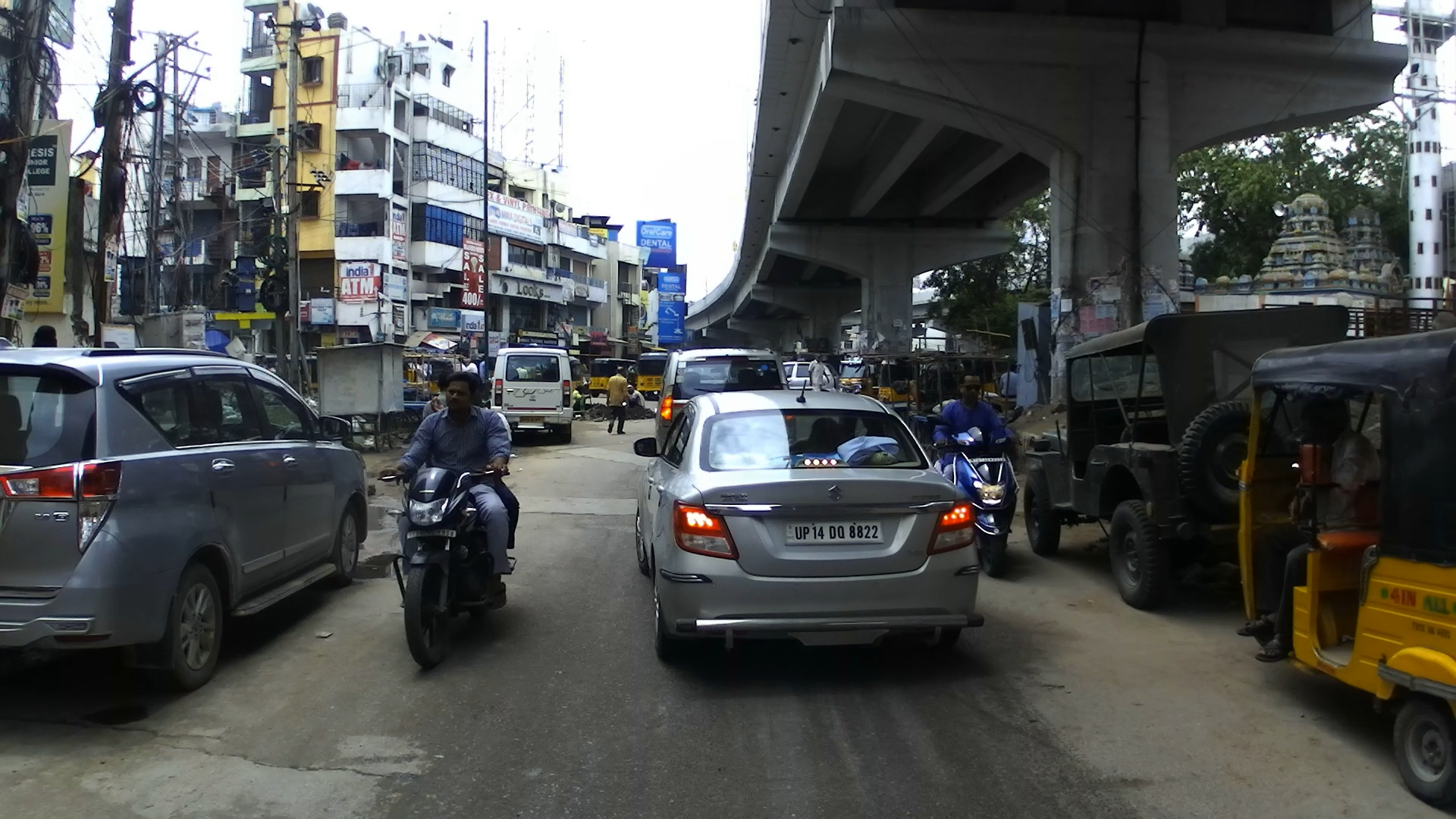}
    \end{subfigure}
    \caption{Traffic participants and driving behavior. Clockwise from top-left: diversity in participants, a pedestrian (bottom-left) jaywalking during heavy traffic, wrong-side driving and illegal parking on the road, and chaotic unmanned junctions.}
    \label{fig:capture_driving}
\end{figure}

\subsection{Comparison with Other Datasets}
Datasets like  KITTI~\cite{geiger2012we}, Cityscapes~\cite{cordts2016cityscapes}, Argoverse~\cite{chang2019argoverse}, and NuScenes~\cite{caesar2020nuscenes} are captured in developed countries where infrastructure is well-developed and road activity is structured. 
The results obtained from these datasets are often not directly applicable in unstructured road situations prevalent in large parts of the world. 
Verma et al.~\cite{varma2019idd} introduced the India Driving Dataset (IDD), a novel dataset for road scene understanding in unstructured environments to meet these needs. It consists of 10,000 images, with 34 classes collected over 182 drive sequences on roads in various parts of India.
They use a rich label set with a four-level hierarchy to represent the nuances of real driving behaviors in semantic segmentation. 
Our attempt is evaluated on situations similar to that of IDD, but at a much large scale in road length and the number of driving hours. In contrast to the previous attempts in IDD, we do not focus on semantic segmentation or the detection of common road objects. We focus on irregularities on the road surface and traffic infrastructure. Fig. \ref{fig:dataset_summary} demonstrates the diversity in the conditions we have used in the evaluation.

\subsection{Challenges}
Constant wear and tear, heavy rains, and laying of underground cables necessitate routine maintenance of roads.
Economic and regulatory hindrances, lack of maintenance, weak accountability, and unreliable funding have resulted in unconstrained road conditions.
As shown in Figure \ref{fig:capture_road}, rough terrains including muddy, bumpy roads, and defects such as potholes, waterlogs and hazardous road objects, are common. 
The problem is compounded due to varying levels of occlusions, unstructured motion, and cluttered background.
Other road infrastructure components that aid navigation is found in a similar state of poor planning and maintenance.
Lane markings either do not exist or are covered in dust, traffic signs are worn out or are concealed amidst the background clutter, and street lights are obscured by vegetation.

As shown in Figure \ref{fig:capture_driving}, there is also a large variation in the traffic participants and their behavior.
Two and three-wheeler vehicles are very prominent and contribute immensely to the unstructured motion on the road. Stray animals and jaywalking pedestrians complicate this problem even further.
Moreover, unsustainable population and vehicle density in cosmopolitan cities, combined with weak enforcement of traffic rules, has brought about disarray in driving behavior.
Non-adherence to road lanes, traffic congestion, and chaotic unmanned junctions are widely prevalent. 
Additionally, violation of traffic rules such as wrong-side driving, swerving, vehicle overloading, and illegal parking cause further confusion.

\subsection{Mobile Imaging System and Data} \label{subsec:data}
\begin{figure}[t]
    \begin{subfigure}{.2367\textwidth}
      \centering
      \includegraphics[width=\linewidth, height=3cm, ]{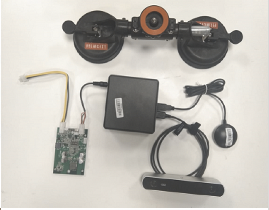}
      \caption{}
    \end{subfigure}
    \begin{subfigure}{.2367\textwidth}
      \centering
      \includegraphics[width=\linewidth, height=3cm, ]{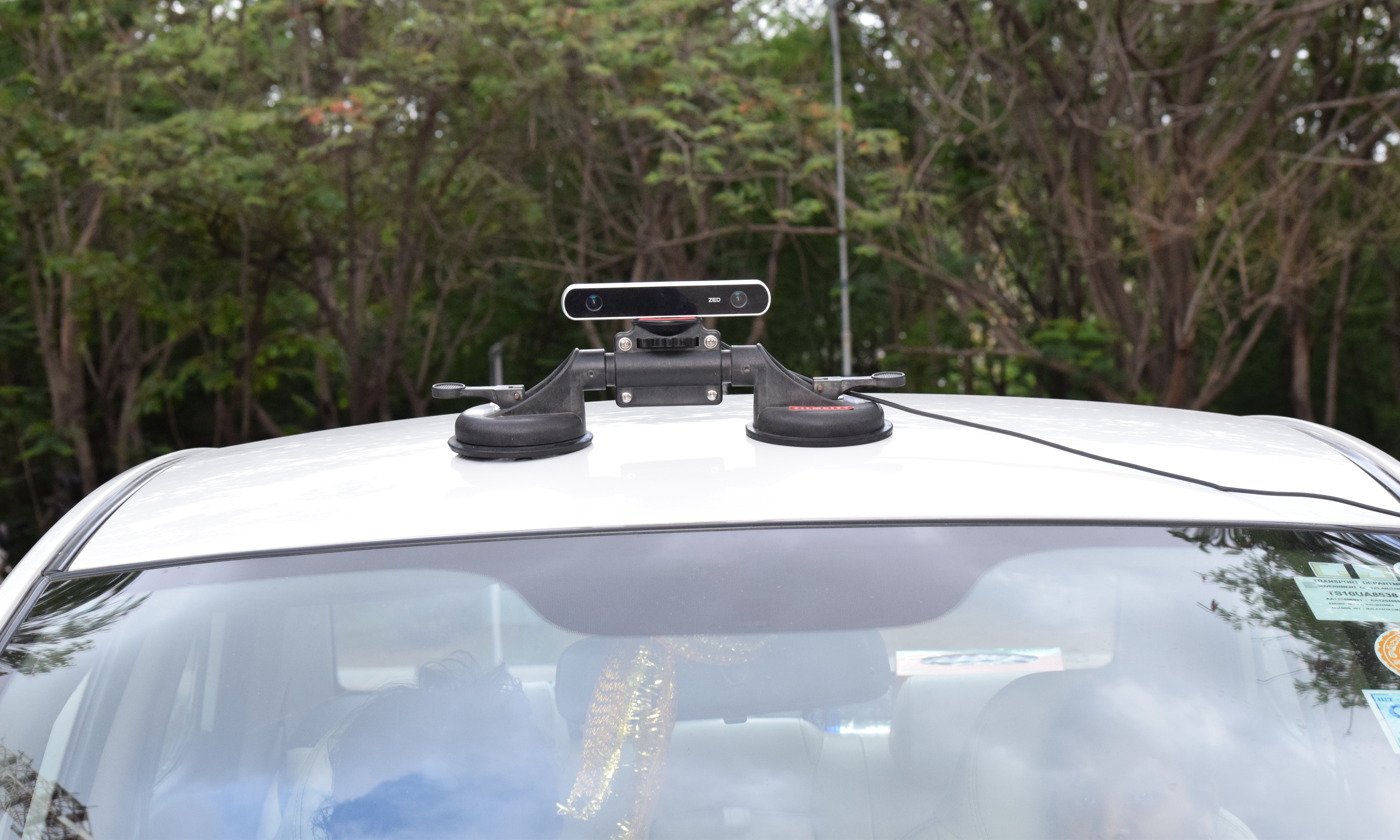}
      \caption{}
    \end{subfigure}
    \caption{Data acquisition: (a) Onboard setup: Camera and GPS connected to Intel NUC unit and suction cup (b) Setup mounted on car used for data capture}
    \label{fig:setup}
\end{figure}

Fig. \ref{fig:setup} shows our setup for data acquisition.
We capture 1920$\times$1080 resolution videos at 15 frames per second.
In total, we record 257 video sequences amounting to 75 hours of capture and covering a distance of over 2000  km.
The sequences have an average duration of 17 minutes and are up to 73 minutes long.
The videos are synchronized to a GPS device polling every second.
The collected videos are annotated at 1-second resolution for road type, traffic density, state of road damage, and time of capture. 
We present this distribution in Fig. \ref{fig:dataset_summary} and the corresponding criterion for classification in Table \ref{tab:data-categories}.

\section{Road Safety Monitoring Pipeline}~\label{sec:system}
\begin{figure}[t]
\begin{center}
\includegraphics[width=8.4cm,height=6cm]{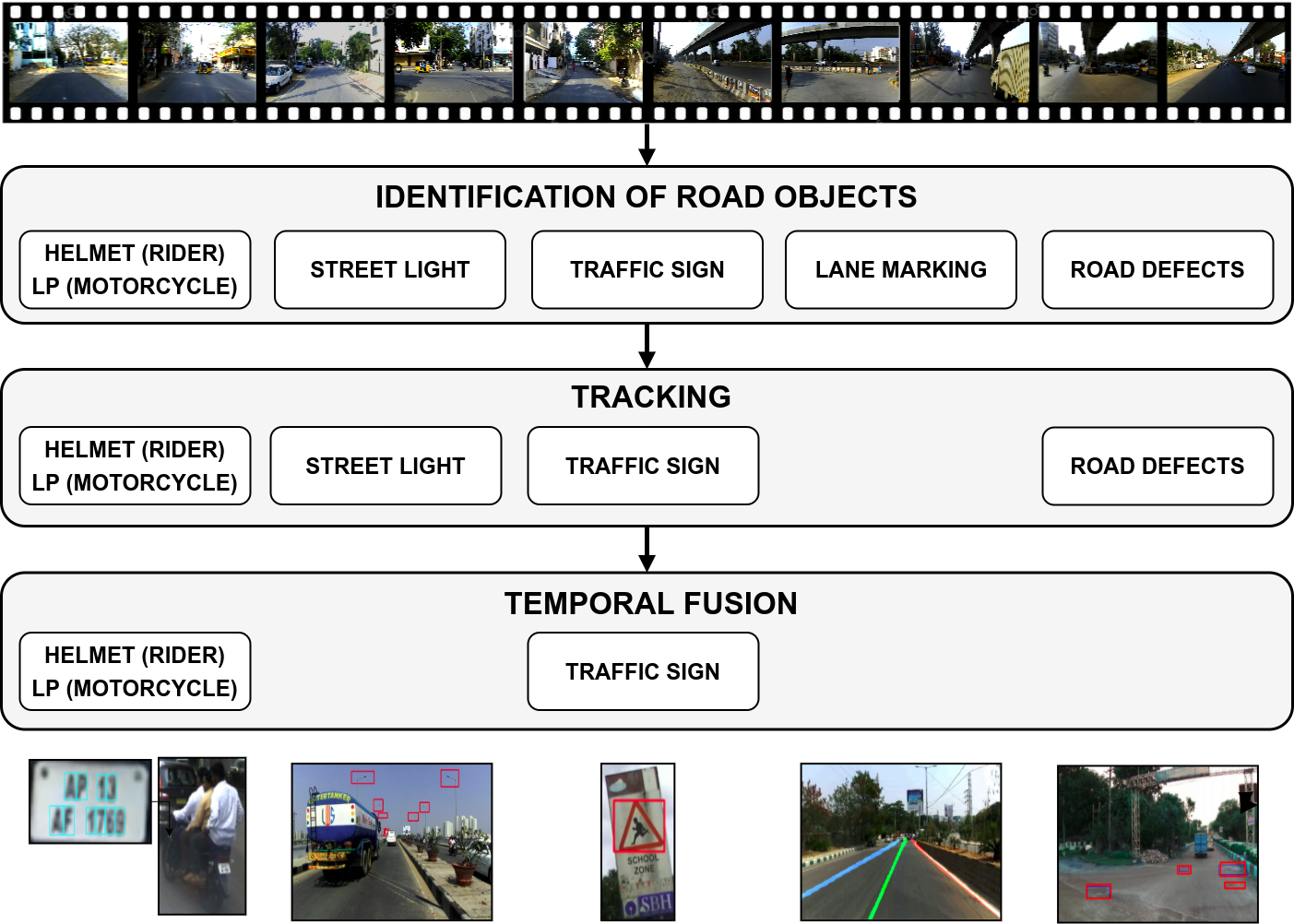}
\end{center}
  \caption{
  Proposed road safety monitoring pipeline. We first detect (or segment) road objects of primary interest in each frame and then track the detected items across the video.
We then leverage temporal viewpoint variance to fuse and strengthen the predictions. }
\label{fig:architecture}
\end{figure}

We are interested in identifying possible irregularities in streets (including   missing lane markings and potholes), absence of street lights, defective traffic signs, and traffic violations from videos.
We achieve this by constructing a simple and efficient pipeline using recent computer vision techniques, as shown in Fig. \ref{fig:architecture}.
In particular, we select computer vision models that achieve (i) high performance and are widely adopted by the community and (ii) fast inference to facilitate evaluation on large data.
We, however, note that the network selection is not a focus of this work and can be replaced with other architectures.

\begin{table}[t]
\begin{center}
\caption{Detection results for several tasks on their respective datasets.
}
\label{table:detection_results}
\resizebox{0.47\textwidth}{!}{%
\begin{tabular}{lcccc}
\toprule

Task / Dataset & Precision & Recall & F1 & mAP@0.5 \\
\midrule
Street Lights & 0.77 & 0.85 & 0.81 & 0.83 \\
Traffic Signs & 0.88 & 0.74 & 0.80 & 0.75 \\
Traffic Participants & 0.82 & 0.79 & 0.80 & 0.81\\
Helmet Violations & 0.78 & 0.83 & 0.81 & 0.83 \\
Potholes & 0.66 & 0.53 & 0.59 & 0.54 \\
\bottomrule
\end{tabular}
}
\end{center}
\end{table}

\subsection{Identification of Road Objects}
\textbf{Lane markings} are continuous and require strong structural priors for segmentation. 
We employ Spatial Convolutional Neural Network (SCNN)~\cite{SCNN} to segment lane markings as it also extrapolates the lane markings occluded by other objects on the road surface. 

\textbf{Traffic signs, street lights, potholes, and traffic participants} are directly detected from the incoming video feed.
We employ YOLOv4 by Bochkovskiy et al.~\cite{bochkovskiy2020yolov4} as it provides a good trade-off between detection performance and speed. 
As part of traffic participants, we detect riders and motorcycles separately and associate them using simple heuristics.

\textbf{Helmet violations} are computed on the detected crops of riders and motorcycles.
We use the WPOD-NET by Silva et al.~\cite{silva2018license}, which detects and geometrically transforms the license plates to a planar front-view. 
For helmet detection on riders, we employ YOLOv4 for the same reasoning mentioned earlier.

\begin{table*}[!ht]
\centering
\caption{mAP scores of our system under varying conditions of road, traffic and environment}
\label{tab:Conditional_evaluation_time}
\resizebox{\textwidth}{!}{%
\begin{tabular}{|l|c|c|c|c|c|c|c|c|c|c|c|c|c|}
\hline
\multicolumn{1}{|c|}{} & \multicolumn{3}{c|}{Time} & \multicolumn{3}{c|}{Traffic Density} & \multicolumn{4}{c|}{Road Type} & \multicolumn{3}{c|}{Road Damage}\\ \hline
Task/Dataset & Morning & Afternoon & Evening & Sparse & Moderate & Dense & Bridge & Narrow & Standard & Highway & Low  & Medium & High\\ \hline 
Street Lights & .93 & .80 & .94 & .90 & .90 & .92 & .89 & .81 & .90 & .82 & .77 & .81 & .91\\ \hline 
Traffic Signs & .50 & .50 & .50 & .54 & .58 & .37 & .33 & .46 & .66 & .44 & .61 & .49 & .46\\ \hline
Traffic Participants & .91 & .87 & .88 & .93 & .85 & .78 & .82 & .98 & .91 & .83 & .85 &  .88 & .86\\ \hline 
Helmet Violations & .66 & .61 & .68 & .72 & .70 & .65 & .72 & .79 & .63 & .65 & .73 & .69 & .82\\ \hline
Potholes & .57 & .46 & .54 & .55 & .52 & .35 & .55 & .42 & .71 & .57 & .56 & .32 & .36\\ \hline 
\hline
Overall & .71 & .65 & .71 & .73 & .71 & .61 & .65 & .67 & .80 & .67 & .70 & .63 & .73\\ \hline 
\end{tabular}%
 }
\caption*{}
\end{table*}

\subsection{Tracking Identified Objects}
The detected objects viz. traffic signs, street lights, potholes, and traffic participants are passed to a tracker.
By using a tracker, we (i) avoid redundant counting, (ii) temporally fuse predictions, and (iii) smoothen detections across the frames.
We use Simple Online and Realtime Tracking (SORT) by Bewley et al.~\cite{Bewley01} since it is extremely fast and provides relatively high accuracy.
SORT tracks objects by employing Kalman filter to handle motion prediction and the Hungarian method for frame-by-frame data association. They use bounding box overlap of detections as the association metric.

\subsection{Temporal Fusion}
Since subsequent frames provide a different viewpoint of the same object to the network, we combine individual frames' predictions and smoothen any aberration.
We use majority voting \cite{lam1997application}, a standard fusion technique, to combine frame-level predictions of traffic sign classification, rider-motorcycle association and helmet classification.

\bigskip

For initial training and quantitative analysis, we annotate $4.3$K images which are sampled randomly from our data capture. 
We use a 80:20 train-test split for this purpose. We report the detection results for all the tasks in Table \ref{table:detection_results}. 
Our entire pipeline runs at {\raise.17ex\hbox{$\scriptstyle\sim$}}10 frames per second, making it feasible to evaluate on large data.


\section{Evaluation under Diverse Conditions}

The road scene around us is visually diverse and constantly in-transition, with highly-dynamic traffic participants riding on infrastructure that varies by locality and maintenance-levels, and environmental conditions.
Studies have shown that such diverse and dynamic settings affect computer vision algorithms \cite{acm_comm}.
However, prior works that assess road safety do not consider this possible impact on their model performance.
This raises concerns regarding the robustness and reliability of computer vision models while deploying them in the real-world.
Therefore, we evaluate our proposed framework under varying conditions of the environment (time), road infrastructure (road damage, road type), and road activity (traffic).

For this evaluation, we use the hierarchical condition-based labels described in Section \ref{subsec:data}. 
We sub-sample 100 images from each sub-category and annotate them with bounding-boxes for the detection components in our pipeline.
We consider two different design choices for sampling images for a category of interest (say, traffic density). 
In the first option, we select and fix the conditions in the other categories (e.g., morning-time, narrow-road, low-road damage) while sampling images in the category of interest (i.e., low, medium, and high-traffic density).
In the second option, we sample from a uniform distribution without considering the labels of other categories.
While the first option allows interpreting results in a conflict-less manner, they would be biased to the exact set of conditions chosen by us.
Contrastingly, while the second option makes it slightly tricky for interpreting results, it captures the natural variation in data and avoids being biased towards a chosen set of conditions.
Thus, we proceed with the latter option and rely additionally on qualitative results to strengthen our reasoning.

\begin{figure} [!ht]
    \begin{subfigure}{.2386\textwidth}
      \centering
      \includegraphics[width=\linewidth, ]{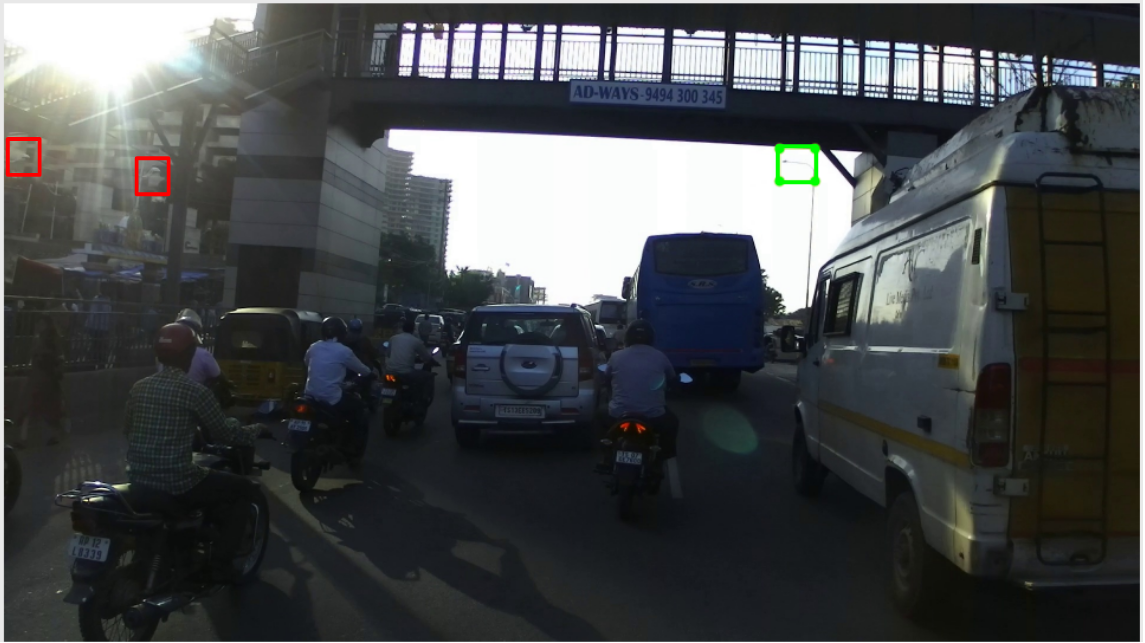}
      \caption{Afternoon time - street light}
      \label{fig:streelight_afternoon}
    \end{subfigure}
    \begin{subfigure}{.2386\textwidth}
      \centering
      \includegraphics[width=\linewidth, ]{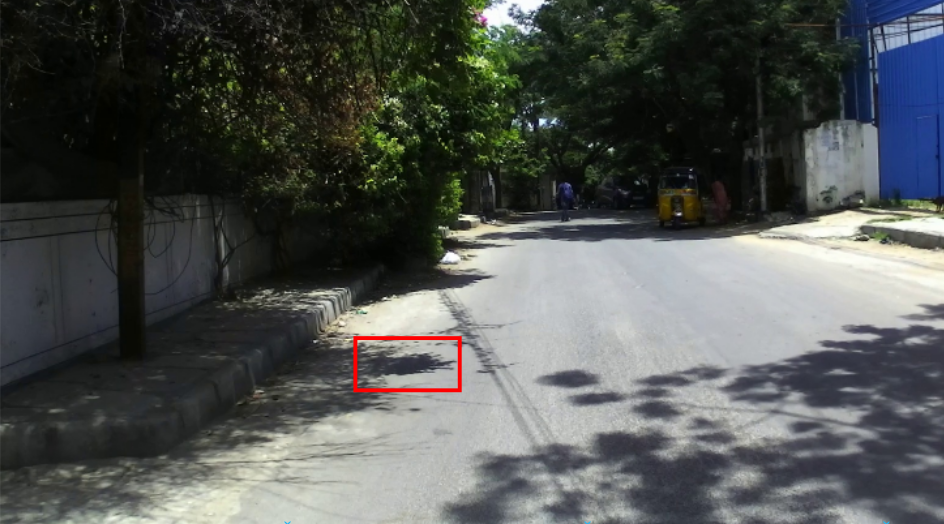}
      \caption{Afternoon time - pothole}
      \label{fig:pothole_shadow}
    \end{subfigure}
    \begin{subfigure}{.2386\textwidth}
      \centering
      \includegraphics[width=\linewidth, , ]{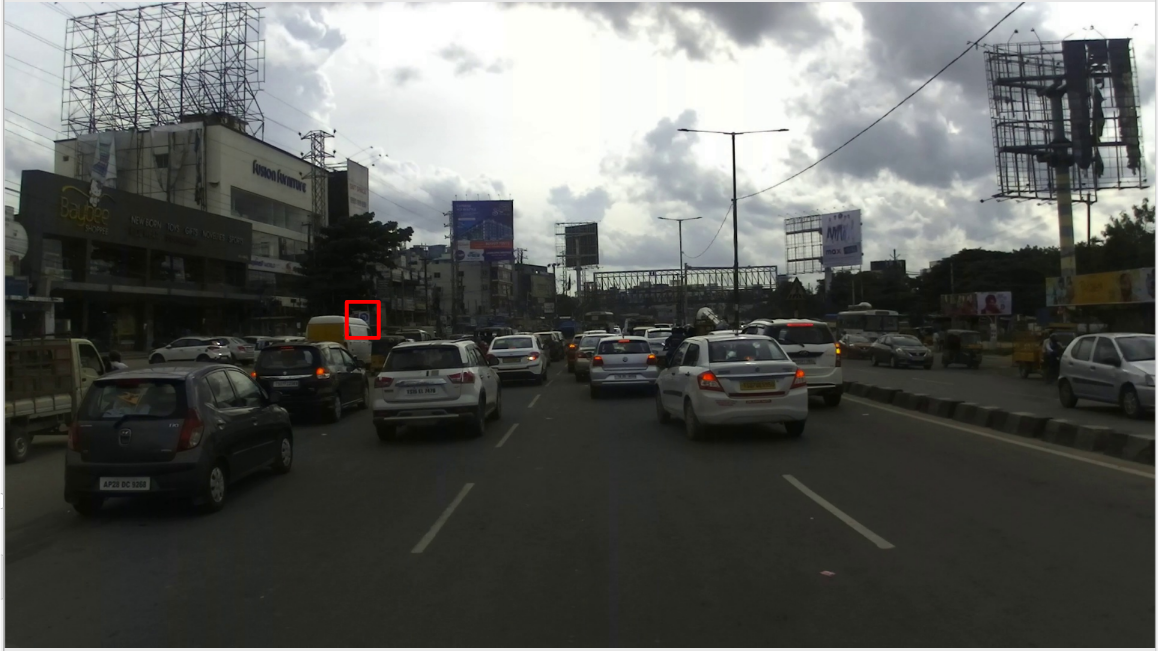}
      \caption{Dense traffic - traffic sign}
      \label{fig:sign_traffic}
    \end{subfigure}
    \begin{subfigure}{.2386\textwidth}
      \centering
      \includegraphics[width=\linewidth, ]{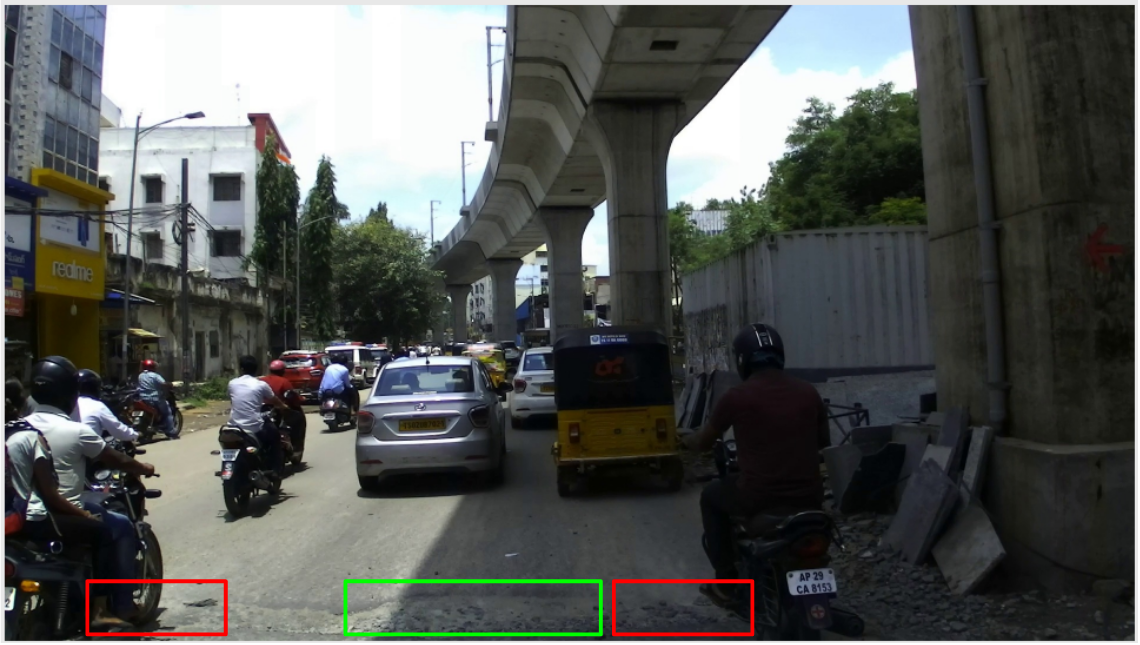}
      \caption{Dense traffic - pothole}
      \label{fig:pothole_traffic}
    \end{subfigure}
    \begin{subfigure}{.2386\textwidth}
      \centering
      \includegraphics[width=\linewidth, ]{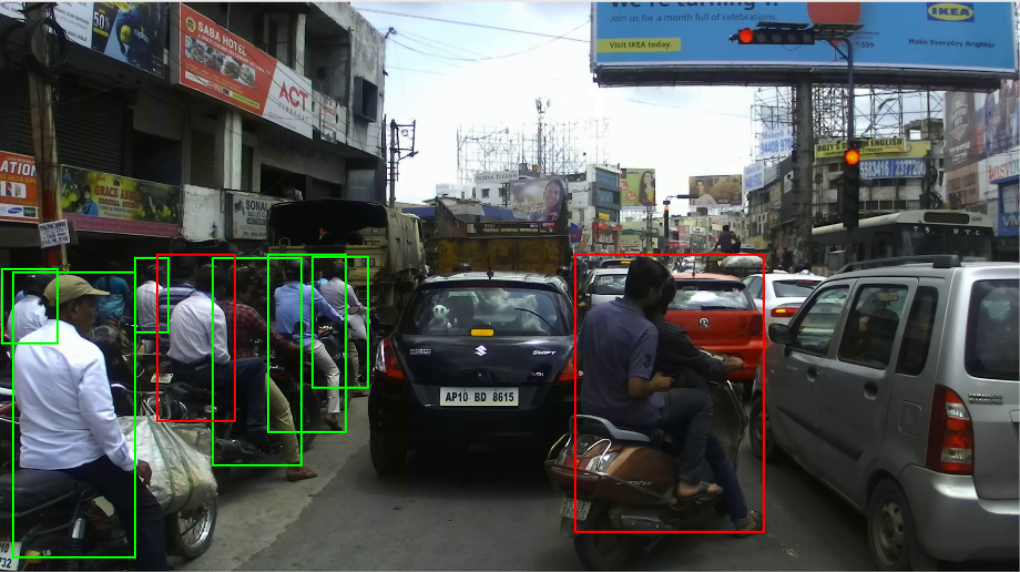}
      \caption{Dense traffic - participants}
      \label{fig:multiple_riders}
    \end{subfigure}
    \begin{subfigure}{.2386\textwidth}
      \centering
      \includegraphics[width=\linewidth, ]{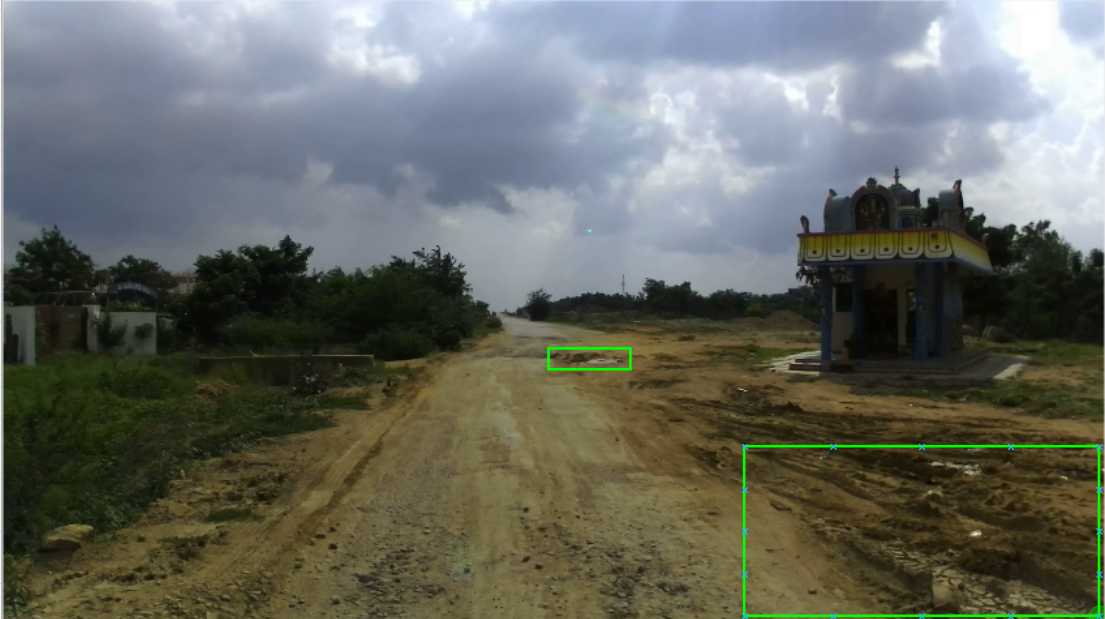}
      \caption{High-damaged road - pothole}
      \label{fig:pothole_muddy}
    \end{subfigure}
    \caption{Predictions under challenging conditions. For each image, only predictions of object of interest are shown. Green and red color boxes indicate correct and wrong predictions respectively. Sub-captions are formatted as (condition - target object).}
    \label{fig:qualitative_results}
\end{figure}

\begin{table*}[!ht]
\begin{center}
\caption{
City-level assessment of various components of road safety. The poor safety scores indicate the need for urgent and significant action.
}
\label{tab:Infra_audit}
\resizebox{.8\linewidth}{!}{%
\begin{tabular}{cccccc}
\toprule
\multicolumn{2}{c}{Traffic signs} & Street lights & Lanes & Potholes & Helmet \\
\cmidrule(r){1-2}
\cmidrule(r){3-3}
\cmidrule(r){4-4}
\cmidrule(r){5-5}
\cmidrule(r){6-6}
Visibility-Range & Defective  & Avg. pair distance & No markings & Percentage of stretches & Percentage of  Violating riders\\ 
\midrule
9.7m & 37.5 \% & 165m & 60.3 \% & 4.0 \% & 45.9 \% \\
\bottomrule
\end{tabular}
}
\end{center}
\end{table*}

We present the results of this experiment in Table~\ref{tab:Conditional_evaluation_time}. 
Analyzing the sub-category values at different times of the day, we notice uniformity in performance, row-wise, while detecting traffic signs, traffic participants, and helmets.
However, there is a noticeable drop in mAP scores of street light and pothole detection during the afternoon session. 
We observe that this is due to false-negatives in the case of street lights and false-positives with regards to potholes.
We attribute the false-negatives to the sunlight's glare on the street lights, whose intensity is at its peak in the afternoon (Fig.~\ref{fig:streelight_afternoon}).
Interestingly, this glare does not affect the performance of other detection modules.
On the other hand, the incidence of the sun's rays is orthogonal to the objects on the road at this time. This produces shadows that are visually similar to potholes and leads to false positives (Fig.~\ref{fig:pothole_shadow}). However, we filter out such false-positive cases by removing the tracks covering few ($\leq3$) frames.

We now look at the results under the traffic density category.
Since street lights are situated much above the road surface, their visibility is not affected by the traffic participants.
However, occlusion arising from dense traffic plays a significant role while detecting traffic signs, traffic participants, and potholes.
Since street signs are mounted around vehicle-height, they are severely impacted by heavy traffic activity on the ground (Fig. \ref{fig:sign_traffic}).
In the case of potholes, traffic participants and their shadows significantly reduce the potholes' visibility (Fig. \ref{fig:pothole_traffic}).
As for traffic participants, we observe that, often, nearby riders are confused as a single entity by the model, leading to false negatives (Fig. \ref{fig:multiple_riders}).

While looking at the state of road damage, we are mainly interested in the detection of potholes.
From Table \ref{tab:Conditional_evaluation_time}, we notice a drop in mAP scores when the damage to the road is medium or high.
However, while inspecting the qualitative outputs, we observe the model outputs to cover most potholes even when the road is highly damaged (Fig. \ref{fig:pothole_muddy}).
We notice that the discrepancy between quantitative and qualitative outputs is due to nearby potholes being detected as a single entity.

To summarize, we notice that the performance of models is acceptable in most settings.
We identify the specific (task, condition) combinations for which computer vision models struggle.
We observe occlusion in dense traffic and sunlight in the afternoon as the major factors influencing performance and back our reasoning with qualitative outputs.
Moreover, we find that the impact of these factors are task-specific, highlighting the need and importance of this study.


\section{City-Scale Road Safety Assessment}
We deploy our scalable system in the real-world and assess the safety of roads at a city-scale.
First, we quantitatively measure the overall safety through carefully constructed metrics.
We then show our interactive dashboard for visually inspecting the outputs at an individual and global level. 
This dashboard serves as a platform for users to analyze and initiate action in a time, labor, and cost-efficient manner.
Our demonstrations are carried out on the 2000 km of data captured on unconstrained roads across the city.

\subsection{Quantitative Assessment}
\textbf{Metrics for Evaluation:}
Traffic signs pre-inform riders of impending road scenes, and hence their range of visibility plays a crucial role.
We measure the visibility range of a traffic sign as the distance (calculated using GPS location) traced by the ego-vehicle from the traffic sign's first frame of detection to its last frame of detection. 
Additionally, we also classify traffic signs as defective or normal. 
For lane markings and street lights, we expect them to occur all along the road periodically. 
Therefore, in these cases, the measure of interest is the road stretches without lane markings and street lights.
We tag the detected street lights geographically and calculate the average distance between street lights along the route.  
For lane markings, we use the normalized lane regularity score described in~\cite{RoadInfra_NCVPRIPG}.
This score considers the percentage of pixels identified as lane markings along a road stretch covering $50$ meters. 
Based on this score, we classify the lane markings as fair, faded, or absent.
To measure the quality of the road surface, we calculate the number of road defects for every 100 meters. 
We then classify these road stretches as poor, average, or fair based on the frequency of potholes.
For helmet violations, we calculate the percentage of riders not wearing results since it portrays road participants' compliance level to traffic rules.\\

\textbf{Results:}
In Table~\ref{tab:Infra_audit}, we present the quantitative measure of road safety in the city. 
The results shed light on the lack of maintenance of roads and road objects.
Traffic signs are visible only at a close range of $10$m, and a significant proportion ($37.5\%$) of them are either rusted or faded out. 
Further, on average, only one street light is present every $165$ meters, making night-driving dangerous.
Lane markings are mostly absent or faded ($60.3\%$), especially on side roads and streets. 
Seldom road repairs result in $4.0\%$ of road stretches with potholes. 
Finally, we observe that $45.9\%$ of riders violate helmet rules, indicating low compliance with traffic rules.
These results indicate the need for urgent actions to ensure the safety of traffic participants.

\subsection{Qualitative Inspection}

\begin{figure*}[t]
    \begin{subfigure}{.48\textwidth}
      \centering
      \frame{\includegraphics[width=\linewidth,height=4cm, draft=false]{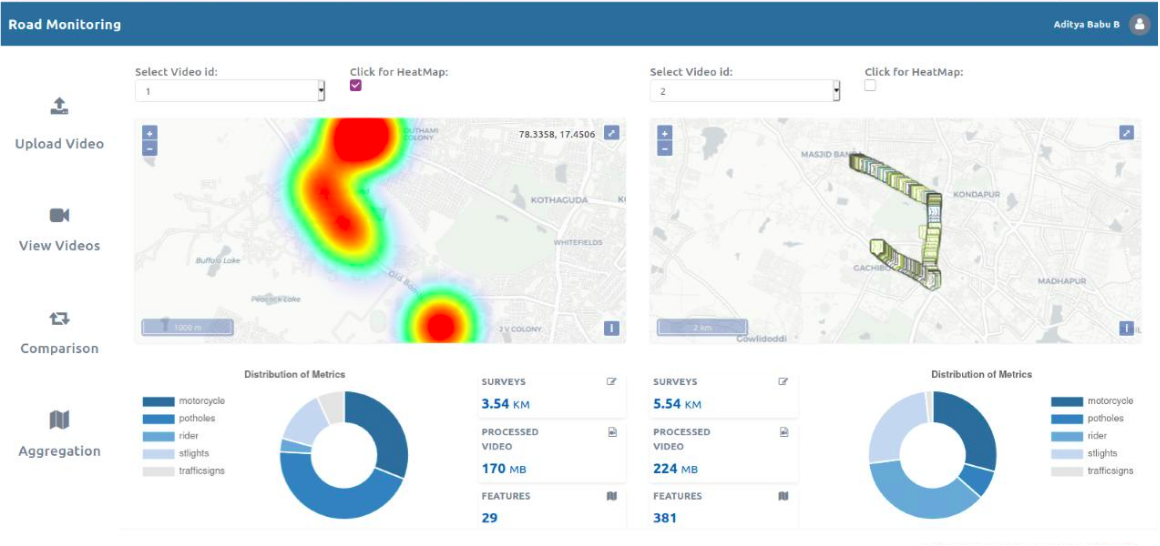}}
      \caption{Powerful qualitative analysis indicating the state of road safety.}
      \label{fig:analysis}
    \end{subfigure}
    \begin{subfigure}{.48\textwidth} 
      \centering
    \frame {\includegraphics[width=\linewidth, height=4cm, draft=false]{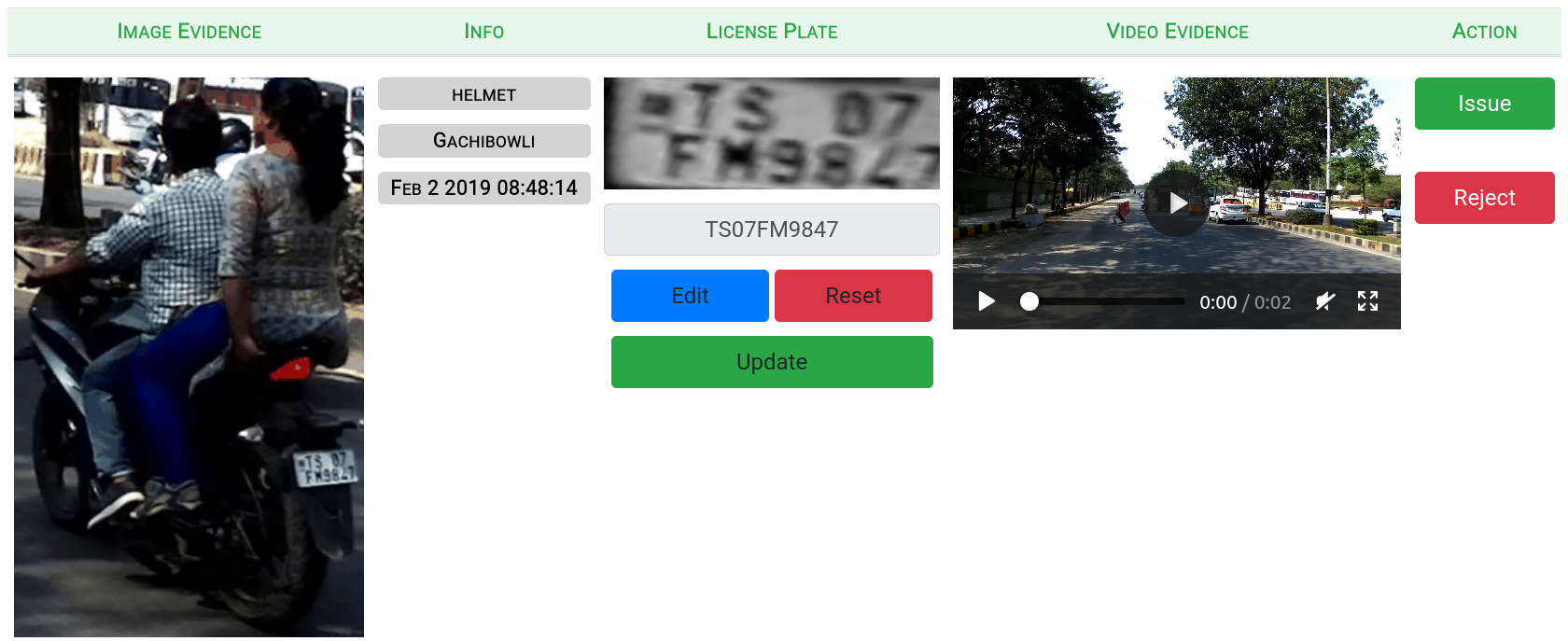}}
      \caption{Interface to verify and issue traffic violation tickets instantly.}
      \label{fig:tickets}
    \end{subfigure} 
    \caption{Our interactive dashboard for easy and efficient visual inspection.}
    \label{fig:dashboard} 
\end{figure*}

Since computer vision models do not work perfectly in the real-world, we develop an interactive dashboard (Fig. \ref{fig:dashboard}) for humans to verify outputs, perform analysis and initiate action.
While we run our framework to identify the objects of interest, we also log the corresponding metadata (time and global coordinates) of capture.
Utilizing this information, we construct heatmaps and route maps that show the state of road infrastructure and compliance-level to traffic rules at a city-scale (Fig.~\ref{fig:traffic_irregularities} (iii) and Fig.~\ref{fig:analysis})\footnote{As discussed earlier, we filter out the false positives (second image from left in Fig.~\ref{fig:traffic_irregularities} (iv) and Fig.~\ref{fig:pothole_shadow})  by removing the tracks covering $\leq3$ frames.}.
Such global visualizations can also help in instantly narrowing down on hotspots in the city, where road safety is in a critical state.
The dashboard provides functionality to zoom-in on these hotspots and inspect the individual outputs, along with image and video evidence.
We can also configure the system to automatically provide warnings when the safety-level falls below a certain threshold.
We provide an additional interface for inspecting traffic violations that makes it effortless to issue tickets after reviewing the image and video evidence (Fig.~\ref{fig:tickets}).
Upon verification, the automatically recognized license-plate number is searched on the vehicle registration database, and a ticket can be issued to the vehicle owner instantly.
We believe this dashboard would serve as a powerful tool to analyze outputs and efficiently initiate action.



\section{Conclusion}
In this paper, we propose a simple mobile imaging setup to address a number of common problems in road safety at scale.
We use recent computer vision techniques to identify possible irregularities on roads, the absence of street lights, defective traffic signs.
Besides inspecting road infrastructure, we also spot traffic violations.
We then investigate the strengths and shortcomings of computer vision techniques on thirteen condition-based hierarchical labels for different timings, road type, traffic density, and state of road damage.
We demonstrate our system on 2000 km of unconstrained road scenes and  quantitatively measure the overall safety of roads in the city.
We also show an interactive dashboard for visually inspecting and initiating action in a time, labour and cost-efficient manner.
We release the trained models, code, and annotations to encourage future work in this direction.

{\small
\bibliographystyle{IEEEtran}
\bibliography{bibliography}
}

\end{document}